\documentclass{article}
\usepackage{arxiv}
\usepackage{nicefrac}       % compact symbols for 1/2, etc.
\usepackage{microtype}      % microtypography
\usepackage{booktabs}       % professional-quality tables
\usepackage[square,sort,comma,numbers]{natbib}
\usepackage{doi}
\usepackage{float}
\usepackage[T1]{fontenc}
\usepackage {graphicx}
\usepackage[english]{babel}
\usepackage[utf8]{inputenc}
\usepackage{academicons}
\newcommand{\orcid}[1]{\href{https://orcid.org/#1}{\includegraphics[width=8pt]{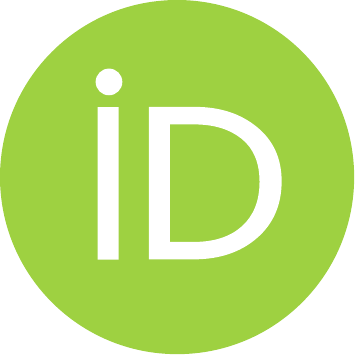}}}

\usepackage{algorithm}
\usepackage{algpseudocode}
\usepackage{xcolor}
\usepackage{float}
\usepackage{hyperref}

\usepackage{amsthm}
\usepackage{subfig}
\theoremstyle{definition}
\newtheorem{definition}{Definition}[section]
\newtheorem{theorem}{Theorem}

\algdef{SE}[SUBALG]{Indent}{EndIndent}{}{\algorithmicend\ }%
\algtext*{Indent}
\algtext*{EndIndent}

\RequirePackage{amsmath, relsize}
\newcommand{\bmdef}{\buildrel{\smaller\triangle} \over =}
\newcommand{\bmparen}[1]{\left({#1}\right)}

\makeatletter
\g@addto@macro\@floatboxreset\centering
\makeatother
 
\pdfoutput=1

\title{Pattern Recognition and Event Detection on IoT Data-streams}

\author{
 Christos Karras\textsuperscript{1}\orcid{0000-0002-4253-7661}, Aristeidis Karras\textsuperscript{1}\orcid{0000-0002-4632-6511}, Spyros Sioutas\textsuperscript{1}\orcid{0000-0003-1825-5565}
\\ \\
\textsuperscript{1}{Computer Engineering and Informatics Department}
\\University of Patras, Patras, Greece
\\
\texttt{{\{c.karras, akarras, sioutas\}@ceid.upatras.gr}}
\\ \\
}

\date{}

\renewcommand{\headeright}{A preprint}
\renewcommand{\undertitle}{A preprint}
\renewcommand{\shorttitle}{Pattern Recognition and Event Detection on IoT Data-streams}

\cleardoublepage
\begin{document}

\maketitle 
\begin{abstract}
Big data streams are possibly one of the most essential underlying notions. However, data streams are often challenging to handle owing to their rapid pace and limited information lifetime. It is difficult to collect and communicate stream samples while storing, transmitting and computing a function across the whole stream or even a large segment of it. In answer to this research issue, many streaming-specific solutions were developed. Stream techniques imply a limited capacity of one or more resources such as computing power and memory, as well as time or accuracy limits. Reservoir sampling algorithms choose and store results that are probabilistically significant. A weighted random sampling approach using a generalised sampling algorithmic framework to detect unique events is the key research goal of this work. Briefly, a gradually developed estimate of the joint stream distribution across all feasible components keeps $k$ stream elements judged representative for the full stream. Once estimate confidence is high, $k$ samples are chosen evenly. The complexity is $\mathrm{O}\bmparen{\min(k,n-k)}$, where $n$ is the number of items inspected. Due to the fact that events are usually considered outliers, it is sufficient to extract element patterns and push them to an alternate version of $k$-means as proposed here. The suggested technique calculates the sum of squared errors (SSE) for each cluster, and this is utilised not only as a measure of convergence, but also as a quantification and an indirect assessment of the element distribution's approximation accuracy. This clustering enables for the detection of outliers in the stream based on their distance from the usual event centroids. The findings reveal that weighted sampling and $res$-means outperform typical approaches for stream event identification. Detected events are shown as knowledge graphs, along with typical clusters of events.
\end{abstract}


\section{Introduction}\label{ch:intro}

\subsection{Internet of Things}

The cutting-edge technology named as the Internet-of-Things seeks to transform the way industries work and behave, as well as the way transportation and production are carried out and manufactured. The term IoT (internet-of-things) stands for the connectivity among devices, sensors and actuators which are connected to each another. It is possible, for example, for authorities in charge of intelligent transportation systems to track and evaluate the vehicle's behaviour and travel after it has left their control, as well as estimate its future location and road traffic flow. The word "interoperable linked things" was initially introduced in 2009 to describe to objects that are uniquely identified and interlinked via the use of radio-frequency identification (RFID) technology \cite{ashton:2009}.

Scientists eventually started connecting IoT to other equipment, such as smart items, GPS devices, and smartphones, in order to enhance its possibilities. Modern definitions of the IoT include a highly globalised communications infrastructure that is self-configurable on the basis of open and coherent communication systems in which actual and virtual 'Things' seem to have physical characteristics,   virtual personalities and distinct identities and communicate intelligently via smart frameworks, all of which are flawlessly integrated into the global network of information \cite{van:2008}.

More specifically, the linking of devices and sensors, RFID-based tags, and communication technologies forms the basis of the Internet of Things. According to \cite{kopetz:2011}, it shows how a range of physical items and devices in our surroundings may be linked to the World Wide Web, allowing these modules and equipment to collaborate and communicate in order to achieve shared objectives.

According to \cite{li:2012} IoT is gaining traction across a broad variety of businesses.
Several industrial IoT initiatives have been completed across numerous sectors, including agricultural, food processing, environmental monitoring, and security surveillance, to name a few examples. Aside from that, the number of publications dedicated to the Internet of Things is rapidly increasing as well.

In addition to being varied and dynamic, IoT ecosystems are penetrating virtually every aspect of daily life, with IoT solutions infiltrating almost every aspect of everyday life. The intelligent industry, which often addresses the expansion of smart production environments and interconnected industrial facilities under the banner of Sector 4.0, is one of the most major application areas. Other areas of importance comprise, but are not limited to, the automotive industry. Smart switches and safety systems are becoming more widespread in the smart housing and construction industries, while smart energy applications highlight the use of intelligent energy, gas, and water metres in residential and commercial buildings. Intelligent transportation solutions such as asset tracking and mobile ticketing are examples, whereas intelligent health solutions such as patient monitoring and chronic disease management are examples of intelligent health solutions. In addition, smart city efforts are looking at technology such as live monitoring of vacant parking spots and intelligent street lighting, among other possibilities \cite{atzori:2010},\cite{fleisch:2010},\cite{vermesan:2014}.

\begin{figure}[htbp]
    \centering
    \includegraphics[scale=0.6]{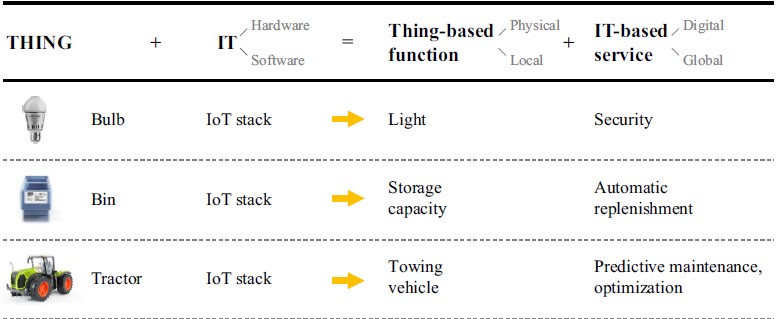}
    \caption{Logic of IoT products and services (source: \cite{wortmann:2015})}
    \label{fig:things}
\end{figure}

To sum it all up, Internet of Things innovation is characterised by the integration of traditional and cyber factors in order to create new products and services. Technological developments in power management, broadband connection, persistent storage, and microchip technology have made it feasible to digitally transform the functions and critical capabilities of products from the industrial era, according to \cite{yoo:2010}. A number of new possibilities for companies to provide value to the Internet of Things have arisen as a result. The value producing notion is illustrated in Figure \ref{fig:things}. In this case, it highlights how commonly Internet of Things solutions are a combination of physical objects and information technology, manifested in the shape of hardware and software.

As depicted, everyday objects can transform to "smart" things with the integration of IoT software and hardware sources enhancing security, replenishment and maintenance.

IoT could well be imagined of as a worldwide architecture of a network made up of various linked devices that use sensing, connectivity, networking, and information processing technologies \cite{tan:2010}. As previously stated, RFID technology is a vital component of IoT, as it allows microchips to wirelessly broadcast identifying information to readers. With the aid of RFID readers, anyone may easily recognise, trace, and manage any item equipped with RFID tags \cite{jia:2012}. Since the 80s, RFID technology has been widely used in various sections including logistics, retail, pharmaceutical production, and product management \cite{sun:2012},\cite{ngai:2008}. However, in our days RFID technology has been substituted by newer and more robust solutions such as LPWAN technologies. These technologies seem to have higher data rate and at the same time lower power consumption \cite{al:2017}.

\subsubsection{Key Features}

IoT is a ground-breaking infrastructure and analytics system which integrates networking, big-data, sensing, as well as AI to construct ultimate solutions targeting a good or service. When applied to an organization or market, these strategies prove to be much more successful. IoT Systems offer a broad variety of applications in sectors due to their flexibility to any environment. These automated systems send and receive data and benefit a variety of activities within an industry. The Internet of Things' essential characteristics include artificial intelligence, device connectivity, sensors, active involvement, and small devices. Specifically:
\begin{itemize}

\item \textbf{Artificial Intelligence}: IoT is inclined to transform everything to "smart," which means that it has the ability to expand every part of our lives via the use of data collecting, algorithms for artificial-intelligence, and networks. This might simply mean that IoT could be in charge of managing the heating and lighting in the house.

\item \textbf{Connectivity}: New network services, primarily for IoT networking, imply that networks are no longer primarily dependent on large providers. Networks may arise on a smaller, more affordable size while still being functional and beneficial. These tiny networks are accompanied by diminutive devices.   

\item \textbf{Sensors/Actuators}: Sensors are critical in the IoT because they operate as determining tools, transforming the ecosystem of devices to interactive system that is capable of actual-world incorporation.

\item \textbf{Active Engagement}: Nowadays, the majority of interactions with linked devices are with passive engagement. The Internet of Things enables direct involvement via the use of a product, a service, or other material.

\item \textbf{Small Devices}: The gadgets have become smaller, cheaper, and more compact over the years. The Internet of Things manages small-scale devices to provide accuracy, scalability, and adaptability.

\end{itemize}

\subsubsection{Benefits}

The benefits of IoT extend to every aspect of life and business. Below, there are a few of the benefits that IoT has:
\begin{itemize}
    \item \textbf{Consolidate Data Collection}: The Modern Data collection in our days is unfortunately limited
    while its design is for passive use. IoT overcomes these limitations and enables humans to analyse the world as they see it. It can be called as an actual image of everything of our world. 
    \item \textbf{Enhancing Customer Engagement}: Unseen faults and substantial defects in the effectiveness of a system scourge data analytics, and as a consequence, involvement stays passive. The Internet of Things totally alters this, allowing for a more robust and productive connection with viewers.
    \item \textbf{Optimization of Technology}: Technologies that enhance the consumer experience also promote device utilization and strive for more significant technological advancements. IoT provides access to a wealth of vital functional and field data.
    \item \textbf{Waste Reduction}: IoT gives real-world data with the least amount of waste, resulting in more efficient resource management.

\end{itemize}

\subsubsection{Limitations}
While IoT offers a plethora of benefits, it also comes with a long list of downsides. The following list discusses the key limitations of IoT.
\begin{itemize}
    \item \textbf{Security}: IoT established an inter-connected devices ecosystem iin which the devices communicate over networks. Despite any security measure, the technology provides minimal control. This exposes users to a variety of threats that attackers can exploit for malicious use.
    \item \textbf{Privacy}: The complexity of IoT enables the collection of significant amounts of personal data in precise detail without the active engagement of the user.
    \item \textbf{Complexity}: According to some, IoT systems are complex to build, install, and maintain due to their reliance on various technologies and a huge number of new supporting technologies.
    \item \textbf{Flexibility}: On the same time, many are worried about an IoT system's ability to interact readily with other systems. They fear that they may end up with a mishmash of incompatible or difficult-to-manage systems.
    \item \textbf{Compliance}: IoT, as any other corporate technology, must adhere to rules. Its intricacy makes compliance seem very difficult, especially since many regard standard software compliance to be a challenge.

\end{itemize}

\subsubsection{IoT Software}

IoT software makes use of frameworks, embedded devices, partner systems, and APIs to handle the critical networking and action issues that IoT ecosystem confronts. These standalone and master apps manage collection of data and interaction, device involvement, real-time insights, application and process extension, and software and process development within the IoT network. Associated operations are carried out via the integration of critical business platforms (e.g., ordering systems, robots, and scheduling). Data Collection: This phase is responsible for the detection, measurement, filtering, and security of light data, as well as the aggregate of that data. It makes use of particular protocols to connect sensors to real-time M2M networks in order to do this. Then it collects data from a large number of sources and distributes it in accordance with the parameters that have been defined. Additionally, it functions in reverse by distributing data across various devices on the network. At some point, the system transmits all of the data it has collected to a centralized cloud server.

In the Internet of Things, device integration software integrates (through interdependent connections) all system devices, resulting in the formation of the system body. It ensures that devices collaborate and communicate in a dependable way at all times. These applications are the software technology that distinguishes the Internet of Things network from other networks; lacking these, the net is not really an IoT system. These take care of the plethora of apps, protocols, and limits that each device has in order to make communication more efficient. Data analytics in real time: These systems collect data or input from a range of sources and translate it into actionable information or identifiable patterns for human analysis. A range of configurations and designs are used to analyze data, which allows them to execute actions connected to automating processes and supplying industry with the information it needs. Utilization and process expansion: These applications broaden the reach of existing software systems in order to build a more comprehensive and efficient system. They bring together certain devices to do specific functions, such as allowing access to particular portable devices or validated tools, among other things. Improved productivity and more exact data collection are two benefits of this technology.

\subsubsection{Common uses}

IoT has applications in every industry and market. It serves a range of user groups, from individuals seeking to minimize their typical household energy use to huge enterprises seeking to simplify their operations. It demonstrates that it is not just beneficial, but virtually indispensable in many sectors as technology progresses and we get closer to the sophisticated automation anticipated in the far future. IoT may be used to improve manufacturing, marketing, service delivery, and safety in various sectors. The Internet of Things enables effective monitoring of several processes, and true openness enables increased visibility of improvement potential. The deep degree of control provided by IoT enables faster and more effective response to such opportunities, which include events such as clear client requirements, nonconforming products, equipment breakdowns, and distribution network difficulties, among others. When used to governance and safety, the Internet of Things enables greater law enforcement, municipal planning, and economic management. The technology reduces existing gaps, by correcting several defects, and enlarges the horizon of these initiatives. For example, in a smart city scenario, a better assess could be done by city planners in which the impact of their designs and governments could potentially be helped by IoT in order to boost the local economy.

In our everyday lives, the Internet of Things delivers a customized experience from our homes to our offices to the businesses with whom we do business with. The result is an improvement in our general well-being, productivity, health, and safety. We can benefit from the Internet of Things in particular when it comes to tailoring our office environment in order to enhance our productivity. Moreover, IoT in Health and Medicine propels us forward toward our envisioned future of medicine, which will be based on a highly integrated network of sophisticated medical equipment. Nowadays, IoT has the potential to significantly improve medical research, equipment, treatment, and emergency care. The integration of these parts results in increased precision, increased attention to detail, quicker response times to events, and continuous development, whilst also lowering the traditional overhead associated with medical research and organizations.

\subsection{Data Streams}

Data stream refers to incoming data that arrives at a very rapid pace. Due to the fact that a high rate indicates a strain on communication and processing infrastructure, based on \cite{muthukrishnan:2005} it can be challenging to:
\begin{itemize}
    \item $trasmit$ (Tr) the whole stream to the program
    \item $compute$ (Co) heavy computationally operations
    \item $store$ (St) absorb a piece or the whole stream immediately 
\end{itemize}

The majority of individuals do not consider this amount of stress when considering Tata Consultancy Services (TCS) capability.
They see data as being contained inside files. When data is communicated, if the connections are poor or the transmission is faulty, we may experience delays, but accurate and full data finally reaches its destination. If the processing capacity is limited or the program is complex, a long time to get the required result will occur, but we will eventually obtain it. Additionally, we store all necessary data. This simplified view of TCS needs is appropriate because, so far, the demand has been matched by available resources as we have generated the volume of data that a technology can transmit, compute, and store.

Two recent events have combined to create significant hurdles for the TCS infrastructure.

\begin{definition}[TCS Approach]\textit{
Capability of automatically generating very comprehensive data streams with continual changes.}
\end{definition}

As is common with real time data, the compiled code is adjusted to reflect the updates, although real-time queries, such as searching up a number, are very uncomplicated. This is also true for purchases with bank credit cards. The most advanced analytics, such as pattern planning and prediction, are often performed offline in warehouses. In contrast, current data streams are generated automatically as a result of sensing devices, which may be linked to the atmosphere, astronomy, networking, finance, or sensors amongst other things, and which generate current data streams. Among their responsibilities are the detection of outliers and extreme events as well as scams, incursions, and unexpected or aberrant behaviour; the monitoring of complicated relationships; the tracking of trends; assisting exploratory research; and the completion of difficult tasks such as classification and signal analysis. Processes that are time-sensitive and must be done in near real-time are required to correctly keep up with the stream's pace, changes, and accurately reflect quickly changing data patterns.

\subsection{Pattern Recognition}

Patterns are a common occurrence in every topic as they depict specific sequences of events that can happen in a regular basis. Pattern recognition is a process in which a ML technique identifies and recognizes regularities (also know as patterns) in data. This technique could be considered as a higher developed data classification. More precisely, PR is considered as a complex process of sequential steps until the desired result is met. These steps include i) data analysis of input data, ii) extraction of possible patterns, iii) comparison among certain standards and iv) use of the results obtained to further assist the PR system. As a result, a pattern recognizer must be capable of performing a diverse range of tasks. Automatic recognition of known patterns is a fundamental skill. However, in many circumstances, for the system to work successfully, it must be able to detect and categorize foreign items, as well as recognize objects even when the data about them is inadequate. The way that the PR algorithms operate is based on three approaches a) the statistical approach which is a statistical decision-theory-based technique. Specifically, the pattern recognizer collects quantitative characteristics from the data and compares them across several samples. It does not, however, discuss how such characteristics are linked to one another. Next, b) is the structural approach which is a more accurate representation of how human vision works. It collects morphological characteristics from a single data sample and examines their connectivity and relationship to one another. And the third approach c) is the neural approach. Specifically, artificial neural networks are used in this strategy. In comparison to the others, it offers for more learning flexibility and is the closest to natural intelligence. 

The essential components of systems for PR are as follows: each method for pattern recognition based on machine learning entails the following phases. In the beginning data entry where n numerous sensors input large volumes of data into the system. Subsequently, segmentation or pre-processing follows where the system organizes the incoming data at this level to prepare the sets for future analysis. Then. the selection of features (or extraction) follows where the system looks for and identifies the unique characteristics of the prepared data sets. In the sequel, classification occurs where the data is classified (or clustered) based on the characteristics found in the preceding stage, or anticipated values are generated (in the case of regression algorithms). Ultimately, post-processing occurs where the system then takes appropriate steps based on the result of the recognition.

Many instances of PR may be found in nature, including individuals recognising faces and dogs responding to their owners' calls, among other things. A broad range of applications, spanning from basic activities to highly specialised areas of knowledge in the field of technology, are based on computer PR algorithms produced via machine learning. Here there are a few common examples:
Virtual assistants, speech-to-text interfaces, and automatic captioning are all examples of NLP (natural language processing) applications; OCR scanners (optical character recognition) applications include mobile scanning apps, medical diagnostic software, and weather prediction software. NIDS (Network intrusion detection systems) are security features that monitor a computer network for patterns of suspicious behavior.

ML algorithms are inextricably linked to pattern recognition algorithms. Unsupervised and supervised learning algorithms are often used when training a PR system to recognize patters in data. As part of supervised machine learning, the human participant provides sample sets of data (referred to as training sets) that are intended to highlight the patterns that the system will be able to identify as it learns. The performance of a system is evaluated once it has processed those data sets, and this is done by exposing it to additional data in a comparable manner, which is arranged into so-called test sets. Classification is the term used to describe the kind of pattern recognition that may be learned in this manner. Because unsupervised machine learning eliminates the need for a human component and eliminates the need for pre-existing patterns, it is the most efficient method of learning. In this example, the algorithm is taught to recognize new patterns without the need to use any previously existing labels, just by being exposed to huge amounts of data over time. In this method, clustering algorithms like as hierarchical or k-means clustering are often utilized. As a result, the pattern recognition acquired via this sort of learning is referred to as clustering or clustering learning. When it comes to neural networks, deep learning is used in conjunction with machine learning to train pattern recognizers, which is a kind of pattern recognition.

The types of pattern recognition are shown in table \ref{tab:typesofpr}.

\begin{table}[htbp]
\centering
    \begin{tabular}{|l|l|}
    \hline
    \textbf{Type}  &  \textbf{Applications}  \cr
    \hline
    Image  &  Visual search  \cr
           &  Surveillance alarm detection  \cr
           &  Face detection  \cr
           &  Optical character recognition  \cr
    \hline
    Text  &  Emotion recognition  \cr
          &  Argument mining  \cr
          &  DNA sequencing  \cr
    \hline
    Sound  &  Mobile or Web applications  \cr
           &  Identifying animal species  \cr
           &  Internet of things  \cr 
           &  Auto captioning for videos  \cr
           &  Melody recognition  \cr
    \hline
    Speech   &  Emotion detection  \cr
             &  Virtual assistants  \cr
             &  Voice-to-text converters  \cr
    \hline
    \end{tabular}
\caption{\label{tab:typesofpr}Types of Pattern Recognition}
\end{table}

PR is now employed as the foundation for a variety of technologies that are commonplace in our daily lives. Face recognition is one of the most prevalent instances of PR at high degree of complexity, as it includes processing of a huge number of visual features that distinguish the face of a person from others. Face Recognition (FR), as well as other biometrics technologies, have already had a significant impact on the process of identity verification, and they will continue to have an impact on our society in the coming years. In addition, pattern recognition is a valuable analytical tool that cannot be replaced. Identifying patterns in large amounts of data is essential for complex big data analysis such as stock market prediction, business analytics, and medical diagnostics. It would be hard to derive meaningful conclusions from vast volumes of data if there were no seamless PR.

\subsection{Event Detection}

When it comes to IoT systems, abnormalities may arise on a regular basis, just as they might in any other informative system. Generally speaking, anomalies may be divided into two categories: 1) anomalies associated with technical or system failure; and 2) abnormalities associated with security breaches (e.g., malicious attacks, penetration to critical components). Anomalies may arise in both categories at the same time on occasion. Hence, a critical stage of an IoT application deployment is to prepare the system for such occasions. These anomalies could be referred as rare events within an applications' normal state. 

%%%%%%%%%%%%%%%% na paei allou %%%%%%%%
In IoT applications, data streams are often comprised by volume and speed. Hence, algorithms should be scalable for in-memory computations. Infeasible batch processing due to data volume Real-time streaming applications often utilize Count-Sketch \cite{charikar:2002}, Bloom \cite{guo:2009} and Bloomier \cite{levenberg:2009} filters, sampling techniques \cite{manku:2002}, and streaming algorithms. 
%%%%%%%%%%%%%%%%%%%%%%%%%%%%%%%%%%%%%%%%
Moreover, event-detection systems shall be developed in such way that the IoT-based ecosystem can successfully identify those occurrences, notify the administrators and wait for action. In a more generic scope, the method of event detection is to analyze event streams in order to uncover collections of events that fit the patterns of events in an event environment. Event patterns and event contexts are used to describe event kinds. If a collection of events matching an event type's pattern is detected during the analysis, the event type's subscribers should be notified. Typically, analysis requires event filtering and consolidation. As expected, event Detection is crucial in almost every IoT deployment from an operational perspective with examples from digital health to smart cities. Ultimately, events and patterns are often considered as outliers in the literature \cite{fawzy:2013}.

\section{Related Work}\label{ch:work}

%\vspace{3cm}
%\begin{quote}
%\textit{``You can't connect the dots looking forward;\\you can only connect them looking backwards.\\So you have to trust that the dots will\\somehow connect in your future.''}\\
%Steve Jobs 
%\end{quote}
%\vspace{3cm}

\subsection{Overview}
In section \ref{sec:rw:streaming} the existing literature and related work are addressed. This section includes an outline in data-mining and pattern mining. The different window models are discussed as well as techniques for sampling and extracting patterns from data-streams. In section \ref{sec:rw:clustering} the preexisting knowledge base and associated work are highlighted covering methods for clustering among $k$-means and more advanced ones specially configured for data-stream clustering. 

\subsection{Streaming}\label{sec:rw:streaming}

Gaining useful insights from streams of data is a difficult topic in data-mining \cite{krempl:2014},\cite{ramirez:2017}, much more so when it comes to pattern mining \cite{gaber:2005},\cite{jiang:2006}. The majority of the existing methods focus on extracting each and every of the frequent occurrences and more infrequently, are restricted only to the top-$n$ most often occurring patterns \cite{wong:2006} or another metric such as max-frequency \cite{calders:2014}. Item-sets is the most often used pattern language, and only a few research have concentrated on more specialized versions, such as patterns that are maximal \cite{karim:2018} or closed \cite{martin:2020}. There are numerous types of window models \cite{jin:2007} which are built to take into account: a) the complete the landmark window where the stream from a certain time is taken \cite{chi:2004,manku:2002,wong:2006}, b) the sliding window where merely the data observations contained inside a window are taken \cite{chi:2004,martin:2020,tanbeer:2009} or c) the damped window where the whole data stream by weighing observations in such a way that the more recent ones are favored  \cite{giannella:2003,raissi:2007}. Undoubtedly, the latter is the more complicated model and at the same time the one that receives the least attention in the existing literature. The plurality of approaches focus on data structure similar to trees for storage and modification of the hitherto mined patterns efficiently and effectively \cite{giannella:2003,manku:2002,martin:2020,tanbeer:2009}. This structure is continuously updated in order to sustain a collection that takes the selected window into consideration. This structure is updated it real-time to keep track of a collection that considers the specific selected window. The Chernoff bound \cite{wong:2006} or similar statistical approaches are frequently utilized to quantify the prevalence of suitable patterns to safely exclude less appealing ones. 

As opposed to progressively collecting intriguing patterns, another the objective is to incrementally gather a representative sample from the data-stream that benefits from RS \cite{efraimidis:2006,vitter:1985}. The primary objective here is to find trends in the data collection by simulating different window models (e.g., SW \cite{babcock:2001}, EB \cite{aggarwal:2006} or TW \cite{raissi:2007}). However, this strategy necessitates repeating the pattern finding process with each alteration to the data sample, which is quite expensive for both techniques pattern mining and pattern sampling. As a result, it is more beneficial to perform sampling on the output space directly rather than the input space. Existing literature addresses this matter with techniques for sampling sequential data \cite{al:2009} but without sampling patters in data streams. Up to this point, stochastic approaches have included evaluating the indicator $m$ on the whole set of data to identify the random walk's next state. Due to the fact that we do not even have every one of the data measurements in a data stream, this evaluation is challenging.

Furthermore, random procedures involving several steps (multi-step) \cite{boley:2011,diop:2020} offer the benefit of avoiding direct measurement evaluation. They include creating an observation of data proportional to the sum of the usefulness of the patterns contained therein and then drawing a pattern proportional to the utility of the patterns contained within. Regrettably, the necessary normalization factor for generating the transaction will be known only after the stream's conclusion. In brief, all pattern sampling approaches demand complete data access, which in the concept of data stream is incompatible, while it cannot hold all data-observations. 

\subsection{Clustering}\label{sec:rw:clustering}
The clustering matter may be approached in several ways, but in this work our primary objective to focus on is $k$-means which is a more popular name for Lloyds Algorithm \cite{lloyd:1982}. In order to increase streaming data clustering, this specific technique has been chosen due to its simplicity. When it comes to identifying the number of classes for the parameter $k$ to be determined for $k$-median \cite{bradley:1996} or in unlabeled multi-class svms (support vector machines) \cite{xu:2005} and similar clustering methods, the principle of using data distribution may also be employed. Data sets are divided into number $k$ of clusters using the $k$-means algorithm. Then, each point of data is classified into one of these groups, which are based on the $k$ random points that were chosen as initial cluster centroids. It's done again and over until the cluster centers converge to stable positions by repeating this technique. In order to have a better idea of how various factors affect the final clustering results, the procedure is repeated numerous times with different centroids in place. A fixed-size dataset may not be an issue, but when dealing with real-time data, this algorithmic trait might lead to significant processing burden. In addition to adding time to the process, $k$-means convergence to clustering via random restarts might result in lower-quality clusters relying on the data source. $k$-means++ \cite{arthur:2006} intelligently picks the first centroids for the initial cluster centers based on randomised seeding. First, a uniform distribution is used to choose a random centre, but the subsequent centres are picked with probability weighted according to how important they are relative to the total potential they represent. However, the technique has exhibited problems when the stream of data changes within time resulting to misclustering in circumstances where the algorithm considers data-sets that can not be handled in-memory as a stream of data (STREAM \cite{guha:2000}) as they are massive. This problem is addressed by the CluStream technique developed in \cite{aggarwal:2003}. CluStream may utilise a user-specified time period to deliver information on past clusters in the area. The stream cluster problem may be divided into two categories i.e., as online microclusters \& offline macroclusters. Since the number of clusters must be predetermined and defined or manually picked at each phase, human oversight is required \cite{aggarwal:2003}. StreamKM++ is another well-known stream clustering method \cite{ackermann:2012}. $K$-means++ \cite{arthur:2006} is the foundation of this technique. However, the clusters must be determined before the experiment can begin. StreamKM++ constructs and preserves a fundamental set of streaming data in order to represent them. So when the data stream has indeed been analysed, the vital data is aggregated using the k-means++ algorithm. Because of this, StreamKM++ was not built to handle dynamic data streams. StreamKM++ In order to determine how many distinct clusters may be discovered in a given dataset, there are a variety of methods. There has been an experimental investigation done in \cite{chiang:2010} in which they introduced the novel technique i$k$-means, which selects the correct $k$ alongside seven other ways. Cluster recovery and the number of instances to use are determined using the following way in their work: groups are identified according to irregularities that are new within the data using Hartigan's criteria, as described in \cite{hartigan:1975}. It was developed in \cite{cao:2006} with the objective of clustering real-time data while taking into account shifting data patterns and streaming noise. Ultimately, DenStream enables the creation and maintenance of dense micro-clusters in real time. When a clustering need is specified, DBSCAN (e.g. \cite{ester:1996}) is utilised to determine the overall cluster result on atop of the subgroups and sub centroids.

\section{Methodology}\label{ch:methodology}

%\vspace{3cm}
%\begin{quote}
%\textit{``Anytime you want to hear about graph partitioning,\\I will be glad to tell you what I know about graph\\partitioning. It remains a standard problem.\\I think it's an interesting problem, because it\\shows up in a variety of guises in real life.''}\\
%Brian Kernighan
%\end{quote}
%\vspace{3cm}

\subsection{Overview}

The implementation consists of several sequential techniques to meet the desired outcome (which is the identification of patters and unique events). Initially, the data-stream is fed on algorithm \ref{alg:weighted}, subsequently algorithm \ref{alg:weighted:random} processes the stream and ensures a random sampling technique to perform unified sampling on the whole stream. Then, the data is fed on algorithm \ref{alg:weighted:random:res} and a reservoir sampling method occurs. Finally, algorithm \ref{alg:kmeans:streams} fetches the $S$ subset of $k$ elements, performs $k$-means improvised clustering to identify outliers which are by assumption events. In section \ref{sec:kg:rep}, the results of algorithm \ref{alg:kmeans:streams} are represented as knowledge graphs where the entity-relation is depicted, as well as, the events identified.

Ultimately, the notation of this work is summarized in table \ref{tab:notation}.
\begin{table}[htbp]
    \centering
    \begin{tabular}{|l|l|l|}
    \hline
    \textbf{Symbol}  &  \textbf{Meaning}  &  \textbf{First in} \\
    \hline
    $\bmdef$ & Definition or equality by definition & Eq. \eqref{eq:n:samples} \\
    $R_n$ & Number of Reservoirs & Eq. \eqref{eq2}  \\
    $r_i$ & Size of Reservoir & Eq. \eqref{eq2} \\
    $M$ & Memory limit & Eq. \eqref{eq:memory}\\
    $R_i$ & Proportion to $r_i$ & Eq. \eqref{eq:proportion}\\
    $d(x,y)$ & Euclidean distance & Eq. \eqref{milnowski}\\
    \hline
    \end{tabular}
    \caption{Notation of this work}
    \label{tab:notation}
\end{table}

\subsection{Stratified Sampling}

A sub-stream is a collection of data pieces preserved from a source. Sub-streams are obtained from a range of different data sources, each of which may have a broad variety of distributions on its own. Stratified sampling was used to ensure that such sub-streams were adequately represented in the sample. In this case, each sub-stream may represent a layer; if a significant number of sub-streams reflect the exact same data distribution, they may be aggregated to represent a single division to simplify the process.

\begin{definition}[Stratified Sampling]\textit{
This technique collects sub-streams from a variety of data sources for each sub-stream, and for each sub-stream, it executes sampling, which might be simple-random-sampling \cite{lohr:2021} or other sorts of sampling individually. As a consequence, data bits out of each sub-stream can be chosen for inclusion in the study in an appropriate way.}
\end{definition}

As a result of stratified sampling, sampling error is reduced, while precision of the sampling procedure rises. It, on the other hand, can only be used in cases when it can make educated guesses about the characteristics for each and every of the sub-streams for example for the length of every sub-stream. When it comes down to it, this assumption of prior knowledge is implausible.

\subsection{Reservoir Sampling}

Commonly, the use of reservoir sampling aids to overcome the false assumption inherent in stratified sample. If the operator does not have foreknowledge of all other sub-streams, the system is still effective. Consider the following scenario: a system is given with a stream including an unknown amount of data pieces.
\begin{definition}[Reservoir Sampling]\textit{
This strategy maintains an infinite data stream of size $R$ and collects a randomly dispersed sample of (at least) $R$ elements from within the unbounded data stream.}
\end{definition}

\begin{figure}[htbp]
    \centering
    \includegraphics[scale=0.85]{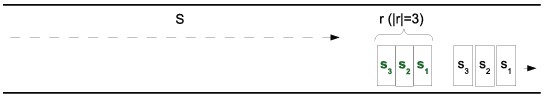}
    \caption{Reservoir Sampling (source: \cite{alkateb:2014})}
    \label{fig:reservoir:sampling}
\end{figure}

This is particularly important because reservoir sampling assures that mostly the earliest $R$ components stay in the reservoir, which is essential. This is accomplished, when the $i$-th item is identified ($I > R$). Reservoir sampling technique keeps it by a likelihood of $N/i$ and then replaces with one of the existing pieces randomly once it is detected. Refer to Figure \ref{fig:reservoir:sampling} for an illustration of how this results in the selection of each item of data from the limitless stream for deposit into the reservoir at random, as indicated. When it comes to resource use, reservoir sampling is successful; nevertheless, the statistical validity of selected data items in a reservoir may be compromised, especially when the arriving data stream is composed of several sub-streams with a diverse range of distributions. It is possible that data from a rarely observed sub-stream would be ignored once reservoir sampling is conducted, for example.

The goal is to distribute $M$ bytes of memory to the reservoirs ($R_{1}$, $R_2$\,\ldots$R_{n}$) of $i$ input streams ($S_{1}$, $S_{2}$,\ldots$S_{\iota}$) so that at the current time time point $t$, the size of each reservoir $r_{i}$ is relative to the number of tuples in total $k_i(t)$ that have arrived so far from $S_{i}$ at the current time point $t$. Equations \eqref{eq:n:samples},\eqref{eq2},\eqref{eq:memory},\eqref{eq:proportion} are obtained from \cite{al-kateb:2007} and they are the foundation for the work implemented in \ref{weighted}.

\begin{equation}\label{eq:n:samples}
n \: \bmdef \:
    \frac{N}{1+N e^{2}}
\end{equation}

where $n$ is the sample size, $N$ is the population size and $e$ is 1-confidence interval.
In our situation, we can find the size $r_i(t)$ of $R_{i}$ by using the formula indicated in the following equation \eqref{eq2}.
\begin{equation}\label{eq2}
r_{i}(t)\: \bmdef \:\frac{k_{i}(t)}{1+k_{i}(t) e^{2}}
\end{equation} 
subject to the given memory limit $M$:
\begin{equation}\label{eq:memory}
\sum_{i=1}^{v} r_{i}(t) \leq M
\end{equation}
We presume that $M$ is ineffectively big to accommodate all of the reservoirs in the system. In this particular scenario, memory is allotted to each reservoir according to the heuristic for memory allocation. As a result of the data-driven computation of $r_i(t)$, the value of $R_i$ is proportional to \eqref{eq2}.

\begin{equation}\label{eq:proportion}
r_{i}^{M}(t)  \: \bmdef \:
    \left\lfloor M\bmparen{\frac{r_{i}(t)}{\sum_{i=1}^{v} r_{i}(t)}} \right\rfloor
\end{equation}

\subsection{Weighted Sampling}\label{weighted}

The issue of random sampling without replacement entails picking $k$ separate random objects from a population of $n$ size and distributing them evenly among them. When all of the objects have the same chance of being chosen, the situation is referred to as homogeneous random sampling. The technique in which items are balanced and the likelihood of each and every item being picked is dictated by the relative weight of the items in the sample is named Weighted random sampling.
Based on \cite{efraimidis:2006}, there really is no such solution as the one that is presented in this work.

\subsubsection{Proof and Implementation}
\begin{theorem}
Assume that $w_i$ represents the weights for the integers $i \in[1,\ldots,n]$. Weighted random sampling without replacement may be defined as algorithm \ref{alg:weighted}.
\end{theorem}

\begin{algorithm}
\caption{Weighted Sampling}\label{alg:weighted}
\begin{algorithmic}[1]
\Require A population comprising $n$ weighted objects in the form of a $V$ population.
\Ensure A subset $S$ of the form $V$ of size $k$.
\State For $i=1$ to $k$
\State $\quad$ let $\pi_{j}(i)=\frac{w_{j}}{\sum_{s_{m} \in V-S} w_{m}}$ be the probability of picking the $j$-th item of $V-S$ in the $i$-th round
\State Select an item $v_{j}$ at random from the $V-S$ subset and place it into the $S$.
\end{algorithmic}
\end{algorithm}

Another approach is shown in algorithm \ref{alg:weighted:random}.

\begin{algorithm}
\caption{Weighted Random Sampling}\label{alg:weighted:random}
\begin{algorithmic}[1]
\Require A population comprising $n$ weighted objects in the form of a $V$ population.
    \Ensure A subset $S$ of the form $V$ of size $k$.
\State For each $v_{i}$, $x_{i}$ is a sample from an independent uniform distribution over the interval $[0,1]$, and the key for each $v_{i}$ in $V$ is defined as $k_{i}=x_{i}^{1 / w_{i}}$
\State $S=$ a subset of the $k$ items with the immense keys
\end{algorithmic}
\end{algorithm}

Algorithm \ref{alg:weighted:random} carries out weighted random sampling without replacement in accordance with the methodology of algorithm \ref{alg:weighted}.

\begin{proof}
Consider first the situation when $k=1$. Calculate the probability of selecting the first item. This requires that the following inequality be satisfied: 
\begin{equation}
x_{1}^{1 / w_{1}}>\max \left(\left\{x_{i}^{1 / w_{i}}\right\}_{i=2, \ldots, n}\right)
\end{equation}

Let $p_{i}$ be defined as the probability that the first item is picked whenever $i=\operatorname{argmax}\left(\left\{x_{j}^{1 / w_{j}}\right\}_{j=2, \ldots, n}\right)$. 

Thus, we have :

\begin{equation}\label{eq:pi}
p_{i}  \:=\:
    \int_{0}^{1} \bmparen{1-x^{w_{1} / w_{i}}} x^{\frac{\sum_{i, \hat{1}}^{w_{j}}}{w_{i}}} d x
\end{equation}

Where $\sum_{\hat{i}, \hat{1}} w_{j}$ is the summation of all weights except the $i$-th and the first items' weights. Thus, it follows that $p=\sum_{\hat{1}} p_{i}$. For convinience, the sums are over all indicies when those are not explicited.

The equation \eqref{eq:pi} is deduced as in the following process: let $x$ be a fixed value for the sample $x_{i}$, then we would have $x_{1}^{1 / w_{1}}>x^{1 / w_{i}}$, which happens with probability $1-x^{w_{1} / w_{i}}$ and $x_{j}^{1 / w_{j}}<x^{1 / w_{i}}$, which happens with probability $x^{w_{j} / w_{i}}$, for all $j \in[2, \ldots, \hat{i}, \ldots, n]$. Then, we multiply all these probabilities (since the samples are drawn independently) and integrate over all the probability space of $x_{i}$ (i.e. interval $[0,1]$ and constant density function equals $= 1$).

By solving the integral \eqref{eq:pi} we find that:
\begin{equation}
p_{i}  \:=\:
    \frac{1}{1+\frac{\sum_{i, 1} w_{j}}{w_{i}}}-\frac{1}{1+\frac{\sum_{i} w_{j}}{w_{i}}}
\end{equation}

\begin{equation}
p_{i}  \:=\:
    \dfrac{w_{i}}{\sum_{\hat{1}} w_{j}} - \dfrac{w_{i}}{\sum w_{j}}
\end{equation}

Thus, we finally find that:
\begin{equation}
\begin{gathered}
p=\sum_{\hat{1}} p_{i}=\sum_{\hat{1}}\left(\frac{w_{i}}{\sum_{\hat{1}} w_{j}}-\frac{w_{i}}{\sum w_{j}}\right)= \\
\frac{\sum_{\hat{1}} w_{i}}{\sum_{\hat{1}} w_{j}}-\frac{\sum_{\hat{1}} w_{i}}{\sum w_{j}}=1-\frac{\sum_{\hat{1}} w_{i}}{\sum w_{j}}=\frac{w_{i}}{\sum w_{j}}
\end{gathered}
\end{equation}
This result corresponds directly to the second phase of the algorithm \ref{alg:weighted}, given that the item with the greatest key in $V-S$ will be chosen in each round.
\end{proof}

The reservoir-sampling version of algorithm \ref{alg:weighted:random} is shown in \ref{alg:weighted:random:res}.

\begin{algorithm}
\caption{Weighted Reservoir Sampling}\label{alg:weighted:random:res}
\begin{algorithmic}[1]
\Require A population comprising $n$ weighted objects in the form of a $V$ population.
\Ensure A subset $S$ of the vector $V$ of size $k$.
\State Initialize the Reservoir $R$ with the first $k$ items of $V$
\State Calculate the key $k_{i}=x_{i}^{1 / w_{i}}$ for each $v_{i} \in R$, where $x_i$ is a random sample from an unbiased uniform distribution across the range $[0,1]$ and $w_i$ is the value of the variable.
\State For each $v_{i}$ in $V-R$ : 
\Indent
    \State let $m$ be the index of the item of $R$ having the smallest key $K$
    \State calculate the key of $v_{i}, k_{i}=x_{i}^{1 / w_{i}}$ (same method) 
    \State if $k_{i}>K$, swap the $i$-th and $m$-th items
\EndIndent   
\end{algorithmic}
\end{algorithm}

The algorithms \ref{alg:weighted:random} and \ref{alg:weighted:random:res} are equal since the algorithm \ref{alg:weighted:random:res} ensures that the items with the greatest $k$ keys will be located within the reservoir, either because they were initially started there or because they will ultimately be exchanged with an item with a lower key value.

The complexity of algorithm \ref{alg:weighted:random:res} in average-case scenario is $O(\min(k,n-k))$.

\subsection{Clustering}\label{sec:clustering}

\subsubsection{Overview}

In the data mining field clustering \cite{drakop:2020:dexa} is a major algorithmic class where a collection of items is to be partitioned based solely on information coming from the items themselves as well as limited external one, typically in the form of hyperparameters. The latter, in contrast to typical parameters, cannot be determined by the data contained in the items and hence have to be tuned either by hand or, as was more recently proposed, special meta-heuristics \cite{bigs:2019:knn}. Perhaps one of the most well-known clustering technique is that of k nearest neighbors (kNN) \cite{bigs:2016:knn} which lends itself to efficient scalable implementations \cite{bigs:2016:kdann+}. Another fundamental subclass of clustering schemes is that of k-means and its numerous variants. One such variant will be presented in this section once the underlying notions are explained.

\begin{figure}[htbp]
\centering
\includegraphics{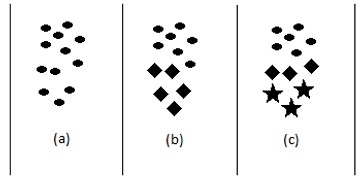}
\caption{Categorizing the data-set in numerous ways (source: \cite{tan:2022})}
\label{fig:cluster}
\end{figure}

Clustering applications may be found in a variety of settings, both in real world and in computer science. When individuals look at items in their environment, they get the impression that they are being classified and grouped depending on their appearance. For example, a farmer who is in charge of classifying fruits according to their sizes throughout the harvesting process. A result of this will be that small fruits will be segregated from large and medium-sized fruits in the categorization system. This is a chore that the vast majority of the people find to be rather uncomplicated. Data mining, in order to do the same thing with objects, takes use of methods for gathering them collectively based on their similarities, which is known to as data clustering. Using classification techniques, it is possible to find similarities between items included within a collection of data by putting them in the same category. According to the \cite{bigs:2016:knn}, the purpose of clustering methods is to identify analogies among elements within a data-set of components. To be more specific, this is accomplished in the $n$-dimensional space by calculating the distance, in which the data dimension is determined by the number of features contained in the collection of characteristics. However, despite the fact that the definition of what forms a group seems to be straightforward, there is subjectivity involved in the process \cite{tan:2022}. Figure \ref{fig:cluster}, for example, demonstrates how different approaches of organising the same data collection of information may be used to get different results. Figure \ref{fig:cluster}(a) depicts the initial data set, figure \ref{fig:cluster}(b) depicts two different clusters, and figure \ref{fig:cluster}(c) illustrates three separated clusters, all of which are clearly depicted in figure \ref{fig:cluster}(a). Figure \ref{fig:cluster}(a) depicts the original data set (a). Even though all of the points in this example are members of the same group in this example, there seem to be methods that cope with group overlaying, which enables a point to be a member of many groups at the same time. In this work, we will only deal with uniform groups, that is, clusters where there is no overlapping between members, as opposed to more flexible groups.

In \cite{rajaraman:2011} the authors describe a number of different data clustering algorithms that are accessible. In this case, items are more alike to the concept of their group than to the models of other groups, and this is the most common pattern in use. An alternative type of grouping is density-based clustering, which gathers items together based on their density in $n$-dimensional space, as opposed to traditional clustering. As a result, things with a high density of occurrences form groupings of objects. Graphs are another kind of clustering technique that is built on nodes (items) that are connected by edges that indicate their relationship.

Existing graph analytics can be applied to knowledge graphs in order to extract latent and non-trivial knowledge from their structure based on connectivity patterns such as degrees, triangles, and higher order paths \cite{drakop:2021:webist}. Such analytics include pattern-based graph resiliency \cite{drakop:2020:ncaa}, deterministic and stochastic graph structural correlation coefficient \cite{drakop:2020:seeda:corr}, directed graph approximation \cite{drakop:2021:adhd}, and graph compression based on two-dimensional discrete cosine transform \cite{drakop:2021:ncaa:graph}. Recently functional analytics have been also developed which, although less generic than their structural counterparts, are tailored to the specific operations taking place on a graph. Examples include graph neural networks operating on affective attributes for ordinary \cite{drakop:2020:adhd} and edge fuzzy graphs \cite{drakop:2021:thecog}, computational aspects of trust \cite{drakop:2020:ictai}, and ontological similarity \cite{drakop:2020:seeda:onto}.

\begin{figure}[htbp]
    \centering
    \includegraphics[scale=1]{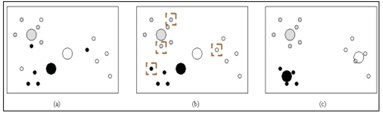}
    \caption{A demonstration of the $k$-means algorithm in action.}
    \label{fig:runkmeans}
\end{figure}

Larose discusses a prototype-based splitting clustering technique known as $k$-means in \cite{larose:2014}. The $k$-means method aims to discover groups by beginning with centroids created by the means of elements inside a cluster. To create groups, a metric of similarity is applied, so that an item seems to be more related to the objects in its cluster than to the items in another data cluster. The $k$-means technique is straightforward, and it has been used in a variety of ways across the scientific world. The estimate user must specify the value $k$, which determines the number of $a$-priori groups we wish to identify using the clustering approach. According to \cite{naldi:2011}, iterative methods are one way for estimating the value of $k$. We outline the steps necessary to construct the conventional $k$-means method, in which the parameter $k$ is a human input parameter, by referring to the aforementioned Algorithm \ref{alg:kmeans}. The figure \ref{fig:runkmeans} illustrates the $k$-means algorithm in action for $k$ = 3.

\begin{algorithm}
\caption{$k$-means algorithm}
\label{alg:kmeans}
\begin{algorithmic}[1]
\Require number of $k$ 
\Ensure clustering on $k$ clusters
\State Select $k$ initial centroids randomly user-defined.
\State Compute the distance between each object's centroids.
\State Allocate each item to the centroid that is closest to it.
\State Calculate the mean for each cluster and use the results to create new placements for the centroids.
\State Recall steps 2, 3, and 4 until the centroid's location remains unchanged.
\end{algorithmic}
\end{algorithm}

In Figure \ref{fig:runkmeans} (a) Each element was randomly allocated to one of three groups and their centroids (bigger circles) were determined. (b) The elements were then reassigned to the groups with the closest centroids. (c) The centroids were recalculated. The groupings have already taken shape. If not, steps (b) and (c) would be repeated until they were.
 
\begin{figure}[htbp]
     \centering
     \includegraphics[scale=0.85]{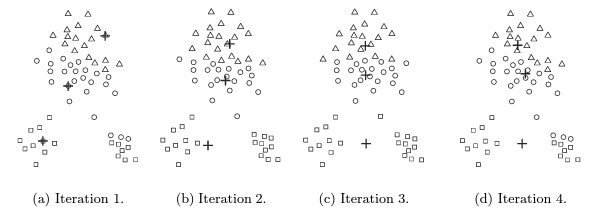}
     \caption{Inadequate initial centroids for $k$-means (source: \cite{steinbach:2005})}
     \label{fig:poor}
 \end{figure}

 However, there are cases that the centroids are selected wrongfully. Such a representation can be found in figure \ref{fig:poor}.
 
 It is possible to estimate the real distance among objects and centroids using closeness relations or similarity metrics, as described in \cite{larose:2014}. The Minkowski measure, which can be determined using the following formula, is one of the most extensively used and quite well metrics in the world.
 
\begin{equation}\label{milnowski}
d(x, y)  \:=\:
    \sqrt[q]{\sum_{i=1}^{t}\bmparen{x_{i}-y_{i}}^{q}}
\end{equation}

This is a Minkowski metric with $q$ = 2 which examines similarities among objects in euclidian space. The Manhattan or (alternatively known as block city distance) \cite{tan:2022} distance is obtained when $q$ = 1.

\begin{table}[htbp]
    \centering
    \resizebox{\textwidth}{!}{%
    \begin{tabular}{|c|c|l|}
    \hline Function of Proximity & Centroid Attribute & Objective Function \\
    \hline Manhattan (L $_{1}$ ) & median & Reduce the sum of an object's L$_{1}$ distance to the cluster centroid \\
    \hline Squared Euclidean (L $_{2}^{2}$ ) & mean & Reduce an object's squared $\mathrm{L}_{2}$ distance from its cluster centroid \\
    \hline cosine & mean & Maximize an object's cosine similarity to the cluster centroid \\
    \hline Bregman divergence & mean & Reduce an object's Bregman divergence to the cluster centroid \\
    \hline
    \end{tabular}}
    \caption{Methods to calculate closeness, centroids, and functions in $k$-means}
    \label{tab:my_label}
\end{table}

\subsubsection{Data Stream Clustering}

\begin{algorithm}
\caption{$k$-means algorithm for Data Streams}\label{alg:kmeans:streams}
\begin{algorithmic}[1]
\Require $k$ number
\Ensure clustering on $k$ clusters
\State When the table of size $x$ is completely filled, or the time $t$ has elapsed, the function gathers data from the stream.
\State Since this is the first cycle, select a beginning medoids at random from a pool of $k$. Begin by taking the cluster centers from the initial state and working your way up to the third iteration.
\State Calculate the distance among each item and its centroids.
\State Allocate every item to the center that is the most closely associated with it.
\State Assume that each group has an average and re-allocate the cluster centers to their new positions.
\State Perform a repetition on Steps 3–5 until the center points do not move.
\State Identify and calculate the clustering SSE for each generated group, as well as the overall clustering SSE.
\State Conserve to a file the ultimate centroid positions, the set of items in each cluster, the SSE for every cluster, and the clustering SSE as a whole.
\State Remove the collected data and return the user to step 1, in which the method is re-performed until the flow is completed.
\end{algorithmic}
\end{algorithm}

With the exception of the conventional phases that are equivalent to the class $k$-means mechanism, the algorithm's first execution chooses the locations of the $k$ centroids relying on the positions of the $k$ arbitrary medoids, such that, on the locations of k database entries, as opposed to the conventional phases that are comparable to class $k$-means methods. As a result of running the $k$-means method, we are able to store the resulting centroids for each cluster as well as the collection of instances, the SSE of each dimension of the different clusters, and the SSE of the overall clustering. As a result, the objects are discarded in order to create place for new ones, allowing for a new clustering process kick-off. A common statistic for analysing clustering approaches is the SSE \cite{tan:2022} statistic. In order to compute the SSE of a group, the equation \eqref{eq:SSE} is used.

\begin{equation}\label{eq:SSE}
\mathrm{SSE}_{\text {group }}  \:=\:
    \sum_{x \in C_{i}} \operatorname{dist}^{2}\left(m_{i}, x\right)
\end{equation}

A point in the group $C_i$ is represented by $x$, while the position of the centroid of the group is represented by $m_i$. Due to the general way clustering works, the SSE of the entire clustering is computed as a sum of all $k$ SSE values detected in the subgroups, as shown in Equation \eqref{eq:totalSSE}.

\begin{equation}\label{eq:totalSSE}
S S E_{\text {total }}=\sum_{j=1}^{k} \sum_{x \in C_{i}} d i s t^{2}\left(m_{i}, x\right)
\end{equation}

Each feature dimension in our programme has its own SSE, which is calculated for each of them. As an example, while working with 2D data, we preserve the SSE for every group, as specified in Equation \eqref{eq:SSE}, in addition to the SSE for each 1D and 2D, as indicated in Equation \eqref{eq:SSE}. The SSE is determined for a single dimension using the equation \eqref{eq:dimension}, as shown in the following equation. The length of a dimension $t$ divided by the total of the squared distances among point $x$ and the cluster center $m_i$ of $t$ dimension. This calculation is performed for every one of the $d$ dimensions of the data set provided.

\begin{equation}
\mathrm{SSE}_{dim}{t} \:=\:
    \sum_{x \in C_{t}} \operatorname{dist}^{2}\left(m_{i_{t}}, x_{t}\right) \quad \forall \quad t \in d
\label{eq:dimension}
\end{equation}

The metric is crucial in determining the overall performance of the clusterings that have been undertaken. Generally speaking, the smaller the standard deviation of a group, the tighter the dispersion of points, and the better the group. The SSE for every dimension is also calculated since it allows us to determine whether or not the data is more evenly distributed in that particular dimension. Aside from the SSE, there are a variety of cluster quality metrics available, including uniformity, group concentration, intragroup peak and lowest values, and plenty of others. Starting with the second cluster of the data stream, the baseline centroids seem to be no longer completely random. The final centroids positions from the last clustering process will be used in this procedure. This inhibits the exchange of positions among two or even more centroids from taking place at the same time. This allows for the visualisation of the evolution of a specific cluster throughout the duration of a data stream, for example.

\subsection{Knowledge Graphs}\label{sec:graphs}

\subsubsection{Overview}

Knowledge graphs are representations that depict information as entities and relationships between them. This kind of modeling of relational information seems to have a long tradition in logic and artificial intelligence \cite{davis:1993}, and is used in a variety of applications such as semantic networks \cite{sowa:2012} and frames \cite{minsky:1974}. The Semantic Data group has more recently adopted it with the intention of establishing a computer-readable "Web-of-data" according to \cite{berners:2001}.

While this notion of the Web 2.0 has not been realised in its entirety, several aspects of these have been accomplished to some extent. Since the W3C Resource Description Framework (RDF) was developed in 2004, the concept of linked data has gained widespread acceptance \cite{berners:2006}, \cite{bizer:2011}. It enables the publication and interrelatedness of information on the Web in relational form through the RDF \cite{klyne:2004}, \cite{cyganiak:2014}.

Primarily, knowledge graphs are composed of datasets from a variety of sources, each of which has a somewhat different structure. Schemas, identities, and context all operate in concert to provide structure to unstructured data. The schemas provide the structure for the knowledge graph, the identities suitably categorise the underlying nodes, and the context defines the context in which the information occurs. These components aid in the differentiation of words with numerous meanings. This enables goods, such as Google's search engine algorithm, to distinguish between the Apple brand and the apple fruit.

Through a process called semantic enrichment, knowledge graphs powered by machine learning generate a complete perspective of nodes, edges, and labels. When data is consumed, knowledge graphs are able to recognize distinct items and comprehend the connections between them. This processability is then evaluated and merged with other relevant and related datasets. Once complete, a knowledge graph enables issue answering and search engines to obtain and reuse exhaustive responses to specific queries. While consumer-facing solutions highlight the system's capacity to save time, the same technologies may also be utilized in a corporate context, obviating the need for manual data gathering and integration to assist business decision-making.

Since knowledge graphs represent evolving fields and by definition can evolve over time through long series of insertions and deletions at random intervals as new data from the underlying domain becomes available, it makes perfect sense to rely on persistent graph structures in order to build and maintain knowledge graphs. Efficient persistent data structures for this purpose among others include ones representing B trees \cite{bigs:2020:btree} and labeled graphs \cite{drakop:2014:ictai}.

Data integration initiatives focused on knowledge graphs may also aid in the generation of new knowledge by building links between previously unconnected data pieces.

\subsubsection{Use Cases}

There are a number of prominent consumer-facing knowledge graphs that are setting the bar for business search systems. Several of these knowledge graphs include the following:
\begin{itemize}

\item  Wikidata and DBPedia are two distinct knowledge graphs for the data on Wikipedia.org. DBPedia takes data from Wikipedia's infoboxes, while Wikidata concentrates on local and global items. Both of these services commonly publish in the RDF format.

\item  The Google Knowledge Graph is accessed through Google Search Engine Results Pages (SERPs), which provide information based on user searches. This knowledge graph has approximately 500 million items and draws its data from sources such as Freebase, Wikipedia, and the CIA World Factbook.

\end{itemize}

Yet, knowledge graphs have applications in numerous of different areas, including the following:
\begin{itemize}

\item \textbf{Retail}:  Knowledge graphs have traditionally been used to support up-sell and cross-sell campaigns by proposing items depending on specific purchase history and popular purchasing patterns across demographic groupings.

\item \textbf{Entertainment}:  KGs are also used to power recommendation engines powered by artificial intelligence (AI) for content platforms such as Netflix, SEO, and social media. These providers propose new material for consumers to read or watch based on their tap as well as other online engagement activities.

\item \textbf{Financial}:  This technology has also been employed in the finance sector for know-your-customer (KYC) and anti-money laundering activities. They aid in the prevention and detection of financial crime by enabling banking institutions to better understand the movement of money among their clients and to identify non-compliant customers.

\item \textbf{Healthcare}:  KGs also aid the healthcare business by facilitating the organization and categorization of linkages in medical research. This data supports doctors in verifying diagnoses and developing treatment regimens that are tailored to each patient's unique requirements.

\item  \textbf{Smart grid}:  The spatial and temporal properties of the electrical power consumption can be employed in order to train efficient consumption predictors \cite{bigs:2021:kou}.

\end{itemize}

%No taxation without representation
\subsubsection{Representation in SPARQL}\label{sec:kg:rep}

\begin{figure}[!htbp]
    \centering
    \includegraphics[scale=0.5]{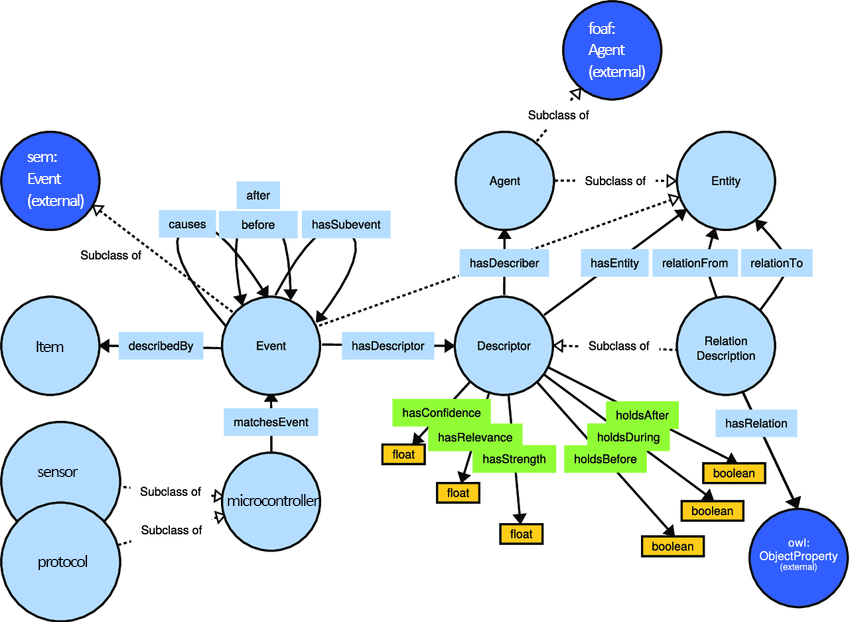}
    \caption{Knowledge Graph Representation}
    \label{fig:kg:sparql}
\end{figure}

\section{Experimental Results}\label{ch:results}

\subsection{Overview}

The experiments for the proposed reservoir sampling method were performed in $Python$ 3.9 language under the development tool $PyCharm$. The packages required for the setup are as follows: $numpy$, $choice$, $torch$. The experiments were conducted with the following hardware: CPU $i9-10850k$, RAM Memory 32GB, Disk Firecuda 530 NVMe and operating system Windows 11. 

In section \ref{sec:uniform-non:samp} the results from uniform sampling with and without replacement are depicted while in section \ref{sec:weighted-non:samp} the results from weighted sampling with or without replacement are depicted. 

In figures \ref{fig:sampling:k2}, \ref{fig:sampling:k8}, \ref{fig:sampling:k2:cuda}, \ref{fig:sampling:k8:cuda} the acronyms w/o stand for with replacement or without replacement.

In tables \ref{tab:k:100:uniform}, \ref{tab:k:4500:uniform}, \ref{tab:k:9000:uniform}, \ref{tab:k:100:weighted}, \ref{tab:k:4500:weighted}, \ref{tab:k:9000:weighted} the acronyms d/r stand for dataset or random data.

\subsection{Uniform Sampling w/o Replacement}\label{sec:uniform-non:samp}

The two evaluation tools used for development and results are CUDA (a parallel computing platform powered by Nvidia) and CPP package of PyCharm (which is a C++ API). The code for the reservoir sampling was implemented in C++ language and the benchmarks are in Python language. The library used to plot the figures is $matplotlib$.
\newpage
\subsubsection{CPP Run}

\begin{figure}[H]
    \centering
    \subfloat[\centering Uniform Distribution w Replacement]{{\includegraphics[scale=0.4]{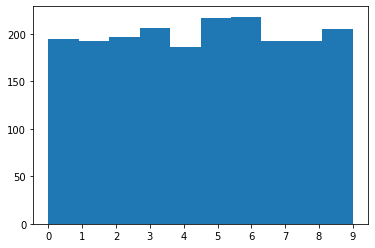}\label{fig:k2:uniform} }}%
    \qquad
    \subfloat[\centering Uniform Distribution w/o Replacement]{{\includegraphics[scale=0.4]{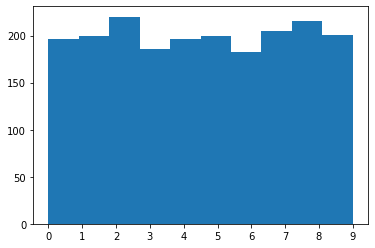}\label{fig:k2:non:uniform} }}%
    \caption{Uniform Sampling w/o Replacement with K=2 on CPP}%
    \label{fig:sampling:k2}%
\end{figure}

For K=2 the results are shown in \ref{fig:sampling:k2}. On \ref{fig:k2:uniform} we have uniform distribution with replacement while on \ref{fig:k2:non:uniform} we have uniform distribution without replacement.

\begin{figure}[H]
    \centering
    \subfloat[\centering Uniform Distribution w Replacement]{{\includegraphics[scale=0.4]{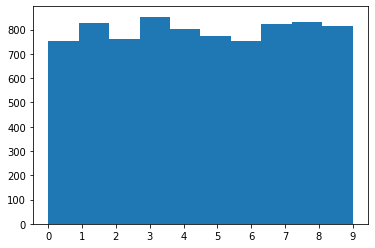}\label{fig:k8:uniform} }}%
    \qquad
    \subfloat[\centering Uniform Distribution w/o Replacement]{{\includegraphics[scale=0.4]{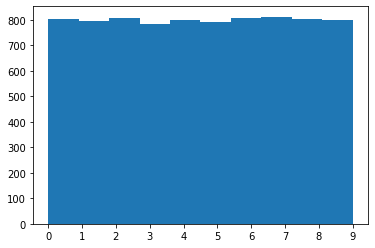}\label{fig:k8:non:uniform} }}%
    \caption{Uniform Sampling w/o Replacement with K=8 on CPP}%
    \label{fig:sampling:k8}%
\end{figure}

For K=8 the results are shown in \ref{fig:sampling:k8}. On \ref{fig:k8:uniform} we have uniform distribution with replacement while on \ref{fig:k8:non:uniform} we have uniform distribution without replacement.

\subsubsection{CUDA Run}

\begin{figure}[H]
    \centering
    \subfloat[\centering Uniform Distribution w Replacement]{{\includegraphics[scale=0.4]{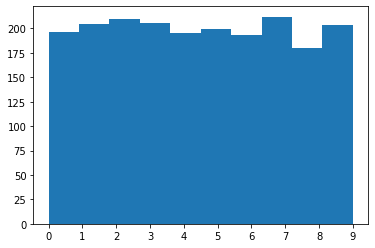}\label{fig:cuda:k2:uniform} }}%
    \qquad
    \subfloat[\centering Uniform Distribution w/o Replacement]{{\includegraphics[scale=0.4]{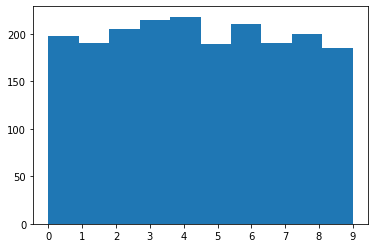}\label{fig:cuda:k2:non:uniform} }}%
    \caption{Uniform Sampling w/o Replacement with K=2 on CUDA}%
    \label{fig:sampling:k2:cuda}%
\end{figure}

For K=2 the results are shown in \ref{fig:sampling:k2:cuda}. On \ref{fig:cuda:k2:uniform} we have uniform distribution with replacement while on \ref{fig:cuda:k2:non:uniform} we have uniform distribution without replacement.

\begin{figure}[H]
    \centering
    \subfloat[\centering Uniform Distribution w Replacement]{{\includegraphics[scale=0.4]{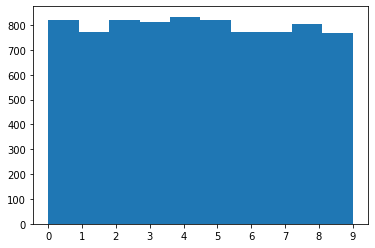}\label{fig:cuda:k8:uniform} }}%
    \qquad
    \subfloat[\centering Uniform Distribution w/o Replacement]{{\includegraphics[scale=0.4]{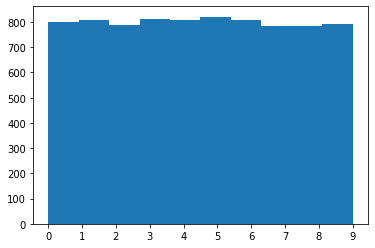}\label{fig:cuda:k8:non:uniform} }}%
    \caption{Uniform Sampling w/o Replacement with K=8 on CUDA}%
    \label{fig:sampling:k8:cuda}%
\end{figure}

\newpage
For K=8 the results are shown in \ref{fig:sampling:k8:cuda}. On \ref{fig:cuda:k8:uniform} we have uniform distribution while on \ref{fig:cuda:k8:non:uniform} we have non-uniform distribution.

The results for bench-marking the uniform w/o replacement for $K=100$ are shown in table \ref{tab:k:100:uniform}.

\begin{table}[H]
    \centering
    \begin{tabular}{|l|l|l|l|l|}
    \hline $K$ & Data (d/r) & Replacement & Time & Mean $+-$ std \\
    \hline 100 & Dataset & Yes & 5.71 $\mu \mathrm{s}$ & 15.9 $\mathrm{~ns}$ per loop \\
    \hline 100 & Random & Yes & 17.3 $\mu \mathrm{s}$ & 73.2 $\mathrm{~ns}$ per loop \\
    \hline 100 & Dataset & No & 14.4 $\mu \mathrm{s}$ & 102 $\mathrm{~ns}$ per loop \\
    \hline 100 & Random & No & 130 $\mu \mathrm{s}$ & 2.07 $\mu \mathrm{s}$ per loop \\
    \hline
    \end{tabular}
    \caption{Benchmarks Uniform Sampling w/o Replacement for $K=100$}
    \label{tab:k:100:uniform}
\end{table}

The results for bench-marking the uniform w/o replacement for $K=4500$ are shown in table \ref{tab:k:4500:uniform}.

\begin{table}[H]
    \centering
    \begin{tabular}{|l|l|l|l|l|}
    \hline $K$ & Data (d/r) & Replacement & Time & Mean $+-$ std \\
    \hline 4500 & Dataset & Yes & 53.2 $\mu \mathrm{s}$ & 1.42 $\mu \mathrm{s}$ per loop \\
    \hline 4500 & Random & Yes & 71.9 $\mu \mathrm{s}$ & 152 $\mathrm{~ns}$ per loop \\
    \hline 4500 & Dataset & No & 72.5 $\mu \mathrm{s}$ & 93.6 ns per loop \\
    \hline 4500 & Random & No & $71.7 \mu \mathrm{s}$ & $258$ ns per loop \\
    \hline
    \end{tabular}
    \caption{Benchmarks Uniform Sampling w/o Replacement for $K=4500$}
    \label{tab:k:4500:uniform}
\end{table}

The results for bench-marking the uniform w/o replacement for $K=9000$ are shown in table \ref{tab:k:9000:uniform}.

\begin{table}[H]
    \centering
    \begin{tabular}{|l|l|l|l|l|}
    \hline $K$ & Data (d/r) & Replacement & Time & Mean $+-$ std \\
    \hline 9000 & Dataset & Yes & $95.4$ $\mu \mathrm{s}$ & $489 \mathrm{~ns}$ per loop \\
    \hline 9000 & Random & Yes & $126$ $\mu \mathrm{s}$ & 1.16 $\mu \mathrm{s}$ per loop \\
    \hline 9000 & Dataset & No & 48.4 $\mu \mathrm{s}$ & 675 $\mathrm{~ns}$ per loop \\
    \hline 9000 & Random & No & 138 $\mu \mathrm{s}$ & 1.42 $\mu \mathrm{s}$ per loop \\
    \hline
    \end{tabular}
    \caption{Benchmarks Uniform Sampling w/o Replacement for $K=9000$}
    \label{tab:k:9000:uniform}
\end{table}

\newpage
\subsection{Weighted w/o Replacement Sampling}\label{sec:weighted-non:samp}

The results for bench-marking the weighted sampling w/o replacement for $K=100$ are shown in table \ref{tab:k:100:weighted}.

\begin{table}[H]
    \centering
    \begin{tabular}{|l|l|l|l|l|}
    \hline $K$ & Data (d/r) & Replacement & Time & Mean $+-$ std \\
    \hline 100 & Dataset & Yes & $43.7$ $\mu \mathrm{s}$ & $536 \mathrm{~ns}$ per loop \\
    \hline 100 & Random & Yes & $110$ $\mu \mathrm{s}$ & 903 $n\mathrm{s}$ per loop \\
    \hline 100 & Dataset & No & 235 $\mu \mathrm{s}$ & 3.67 $\mathrm{~ns}$ per loop \\
    \hline 100 & Random & No & 172 $\mu \mathrm{s}$ & 2.3 $\mu \mathrm{s}$ per loop \\
    \hline
    \end{tabular}
    \caption{Benchmarks Weighted Sampling w/o Replcement for $K=100$}
    \label{tab:k:100:weighted}
\end{table}

The results for bench-marking the weighted sampling w/o replacement for $K=4500$ are shown in table \ref{tab:k:4500:weighted}.

\begin{table}[H]
    \centering
    \begin{tabular}{|l|l|l|l|l|}
    \hline $K$ & Data (d/r) & Replacement & Time & Mean $+-$ std \\
    \hline 4500 & Dataset & Yes & $407$ $\mu \mathrm{s}$ & 4.7 $\mu \mathrm{s}$ per loop \\
    \hline 4500 & Random & Yes & $495$ $\mu \mathrm{s}$ & 8.65 $\mu \mathrm{s}$ per loop \\
    \hline 4500 & Dataset & No & 295 $\mu \mathrm{s}$ & 600 $\mathrm{~ns}$ per loop \\
    \hline 4500 & Random & No & 1.27 $m \mathrm{s}$ & 4.14 $\mu \mathrm{s}$ per loop \\
    \hline
    \end{tabular}
    \caption{Benchmarks Weighted Sampling w/o Replcement for $K=4500$}
    \label{tab:k:4500:weighted}
\end{table}

The results for bench-marking the weighted sampling w/o replacement for $K=9000$ are shown in table \ref{tab:k:9000:weighted}.

\begin{table}[H]
    \centering
    \begin{tabular}{|l|l|l|l|l|}
    \hline $K$ & Data (d/r) & Replacement & Time & Mean $+-$ std \\
    \hline 9000 & Dataset & Yes & $773$ $\mu \mathrm{s}$ & 1.3 $\mu \mathrm{s}$ per loop \\
    \hline 9000 & Random & Yes & $872$ $\mu \mathrm{s}$ & 1.14 $\mu \mathrm{s}$ per loop \\
    \hline 9000 & Dataset & No & 373 $\mu \mathrm{s}$ & 999 $\mathrm{~ns}$ per loop \\
    \hline 9000 & Random & No & 2.92 $m \mathrm{s}$ & 6.24 $\mu \mathrm{s}$ per loop \\
    \hline
    \end{tabular}
    \caption{Benchmarks Weighted Sampling w/o Replcement for $K=9000$}
    \label{tab:k:9000:weighted}
\end{table}

\subsubsection{CPP Run}

\begin{figure}[H]
    \centering
    \subfloat[\centering Weighted Sampling w Replacement]{{\includegraphics[scale=0.4]{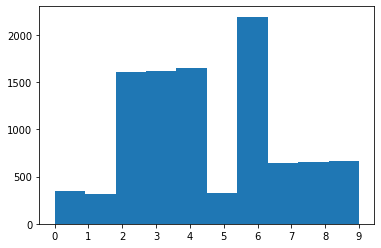}\label{fig:sampling:k1:cpp:weight:true} }}%
    \qquad
    \subfloat[\centering Weighted Sampling w/o Replacement]{{\includegraphics[scale=0.4]{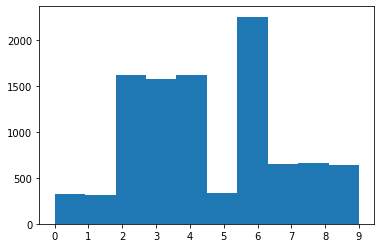}\label{fig:sampling:k1:cpp:weight:false} }}%
    \caption{Weighted Sampling w/o Replacement with K=1 on CPP}%
    \label{fig:sampling:k1:cpp:weight}%
\end{figure}

For K=1 the results are shown in \ref{fig:sampling:k1:cpp:weight}. On \ref{fig:sampling:k1:cpp:weight:true} we have weighted sampling with replacement while on \ref{fig:sampling:k1:cpp:weight:false} we have weighted sampling without replacement.

\newpage
\subsubsection{CUDA Run}

\begin{figure}[H]
    \centering
    \subfloat[\centering Weighted Sampling w Replacement]{{\includegraphics[scale=0.4]{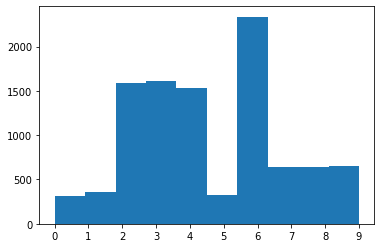}\label{fig:sampling:k1:cuda:weight:true} }}%
    \qquad
    \subfloat[\centering Weighted Sampling w/o Replacement]{{\includegraphics[scale=0.4]{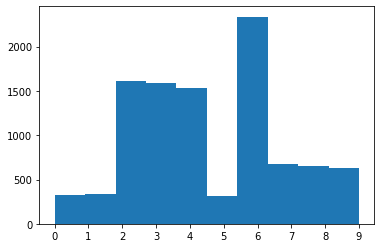}\label{fig:sampling:k1:cuda:weight:false} }}%
    \caption{Weighted Sampling w/o Replacement with K=1 on CUDA}%
    \label{fig:sampling:k1:cuda:weight}%
\end{figure}

For K=1 the results are shown in \ref{fig:sampling:k1:cuda:weight}. On \ref{fig:sampling:k1:cuda:weight:true} we have weighted sampling with replacement while on \ref{fig:sampling:k1:cuda:weight:false} we have weighted sampling without replacement.

\subsection{Proposed Clustering Scheme}
For the training and testing the proposed alternative of $k$-means the SENSORS data-set was used to identify events from IoT sensors. The events detected from the proposed res-means algorithm are shown in table \ref{tab:event:detection}.

\subsubsection{Evaluation}
For the evaluation of the proposed $k$-meas alternative three different methods were compared. The standard $k$-means, DBSCAN and E-DBSCAN. The results are shown in table \ref{tab:eval:comp}.

The evaluation metrics used are precision, recall and accuracy.
$$
\text { Precision }=\frac{\text { True Positive }}{\text { Actual Results }} \text { or } \frac{\text { True Positive }}{\text { True Positive + False Positive }}
$$
$$
\text { Recall }=\frac{\text { True Positive }}{\text { Predicted Results }} \text { or } \frac{\text { True Positive }}{\text { True Positive + False Negative }}
$$

$$
F=2 \times \frac{\text { Precision } \times \text { Recall }}{\text { Precision }+\text { Recall }}
$$

\begin{table}[htbp]
    \centering
    \begin{tabular}{c|c|c|c|c} 
    Performance Measures & K-means & DBSCAN & E-DBSCAN & Res-Means\\
    \hline Precision & $39.7$ & $49.1$ & $53.23$ & 51.2 \\
    \hline Recall & $25.7$ & $40.8$ & $50.1$ & 44.8 \\
    \hline F-Measure & $30.46$ & $44.64$ & $51.45$ & 47.7 
    \end{tabular}
    \caption{Performance comparison of the proposed method}
    \label{tab:eval:comp}
\end{table}

\subsubsection{Event Detection}
The events detected are shown in table \ref{tab:event:detection}. CM, BoW and EBoW are all three similar event-detection frameworks. The different categories have been extracted from the data-set and the proposed method performed clustering across all features. Subsequently, the 4 features are represented as 4 clusters with centroids and outliers. The events shown here are sum of the outliers divided by the distance from the cluster center. Again, we can observe that our method outperforms the other three in all 4 categories.

\begin{table}[H]
    \centering
    \begin{tabular}{l||c|c|c|c} 
Event & CM & BoW & EBoW & Res-means \\
\hline Sensor not working & $4.8$ & $7.1$ & $7.9$ & $\mathbf{8 . 4}$ \\
\hline Noisy data sent & $9.2$ & $9.8$ & $11.2$ & $\mathbf{1 2 . 7}$ \\
\hline Wrong intervals & $4.3$ & $5.2$ & $7.8$ & $\mathbf{9 . 4}$\\
\hline Communication issues & $5.6$ & $6.4$ & $8.4$ & $\mathbf{11 . 2}$
\end{tabular}
    \caption{Events detected with the proposed method}
    \label{tab:event:detection}
\end{table}

\section{Conclusions and Future Work}\label{ch:more}
In this work, a weighted reservoir sampling scheme with complexity $O(min(k, n-k))$ was developed, as well as an alternative approach of $k$-means. The two strategies suggested here compose a system for sampling and identifying patterns and occurrences in large data-streams. These data-streams can be used to gain useful insights from the data arriving from multiple sources. The sampling approach is used to generate a representative subset of the sample $S$ for processing in the micro-controllers' memory. This results in a low cost of calculation and a short computation time. Additionally, it simplifies the $k$-means complexity by providing just a few yet representative data pieces to the algorithm. As a result, the alternative of $k$-means runs swiftly, effectively, and efficiently, calculating distances as well as the SSE for clusters, dimensions, and overall clustering. From a scalability standpoint, the data indicated that both strategies performed satisfactorily. Additionally, the two suggested approaches collaborate and communicate information in the same way they do with low $k$ items as well as with big $k$ items. Noteworthy, is the fact that even with replacement of the $k$ items within the reservoir, the proposed algorithm succeeded better computational time compared to the method where no replacement occurs.  

The work's future directions include improved and more precise reservoir sampling, as well as decreased complexity. Additionally, the system may be employed in distributed IoT ecosystems, where each node (device) can compute and analyse samples obtained from other devices and return them to a central gateway. Another possibility for future enhancement is to include more complicated clustering algorithms to improve accuracy and outlier identification in the suggested clustering strategy. Simultaneous computing of the two described approaches might be a considerable improvement. Thus, the computational cost may be greatly reduced since each clustering technique can be executed immediately after the sampling method is complete, thus lowering the number of I/Os. Moreover, another noteworthy future direction of this work is the implementation of a similar system that takes into account shifts in the distribution of the data. While IoT data comes from many sources, in different intervals and with different rates, it would be significant if we could process these data with a clustering technique that takes into account such limitations.

\bibliographystyle{ieeetr}
\bibliography{reservoir.bib}

\end{document}